\documentclass[letterpaper, 10 pt, conference]{ieeeconf}  
\pdfoutput=1

\IEEEoverridecommandlockouts                              

\usepackage{graphics} 
\usepackage{graphicx}
\usepackage{amsmath}
\usepackage[table]{xcolor}

\title{\LARGE \bf
A Lightweight, High-Extension, Planar 3-Degree-of-Freedom Manipulator Using Pinched Bistable Tapes} 

\author{
O. Godson Osele, Allison M. Okamura, and Brian H. Do
\thanks{This work is supported in part by National Science Foundation grant 2024247 and the National Science Foundation Graduate Research Fellowship Program.}
\thanks{The authors are with the Department of Mechanical Engineering, Stanford University, Stanford, CA 94305 USA
        {\tt\small \{brianhdo, obum, aokamura\}@stanford.edu}}
}

\begin{document}

\maketitle
\thispagestyle{empty}
\pagestyle{empty}

\begin{abstract}
To facilitate sensing and physical interaction in remote and/or constrained environments, high-extension, lightweight robot manipulators are easier to transport and reach substantially further than traditional serial chain manipulators. We propose a novel planar 3-degree-of-freedom manipulator that achieves low weight and high extension through the use of a pair of spooling bistable tapes, commonly used in self-retracting tape measures, which are pinched together to form a reconfigurable revolute joint. The pinching action flattens the tapes to produce a localized bending region, resulting in a revolute joint that can change its orientation by cable tension and its location on the tapes though friction-driven movement of the pinching mechanism.
We present the design, implementation, kinematic modeling, stiffness behavior of the revolute joint, and quasi-static performance of this manipulator. In particular, we demonstrate the ability of the manipulator to reach specified targets in free space, reach a 2D target with various orientations, and maintain an end-effector angle or stationary bending point while changing the other. The long-term goal of this work is to integrate the manipulator with an aerial robot to enable more capable aerial manipulation.
\end{abstract}

\section{Introduction}
High-extension, lightweight robot manipulators have the potential to enable physical interaction in difficult-to-reach environments. Such robots can be more easily delivered than and achieve configurations not feasible for traditional robot arms. One emerging scenario for such manipulators is aerial manipulation -- in which aerial robots physically interact with their environment. Aerial manipulation poses many research challenges, including risk when operating in confined spaces near walls or the ground and strict payload weight constraints, both of which make conventional, bulky industrial robot manipulators prohibitive~\cite{khamseh2018aerial}. Aerial manipulators are typically mounted on the underside of aerial robots, requiring manipulators comprised of revolute joints to undergo complex self-folding into compact designs for stowage when landing~\cite{khamseh2018aerial}. Enabling these manipulators to retract could address this issue, but prior work on retractable arms has focused on arms with just a single degree of freedom~\cite{forte2012impedance, kimOrigamiLock, YANG2019226, zhang2019design}.



In the fields of soft robotics as well as deployable and reconfigurable structures, a number of different designs exist which exhibit retraction capabilities through spooling of material~\cite{hammond2017pneumatic, hawkes2017soft}. One common device that exhibits this capability is the tape measure, with its bistable tape, also sometimes referred to as a tape spring. While many examples exist of bistable reels being used to create deployable structures, with applications ranging from antenna booms~\cite{daton2000deployable} to entertainment products~\cite{honeck2018sword}, these create straight rigid cylinders designed to resist bending. The rigidity of these bistable tapes comes from their transverse curvature, which increases the energetic cost of bending longitudinally and thus dramatically increases the bending stiffness of thin sheets and strips~\cite{taffetani2019limitations, pini2016two}. This curvature-induced rigidity can be observed in the strategy for holding a floppy slice of pizza with one hand or in a tube formed by a rolled-up sheet of paper. 




\begin{figure}[t]
    \centering
    \includegraphics[width=\linewidth]{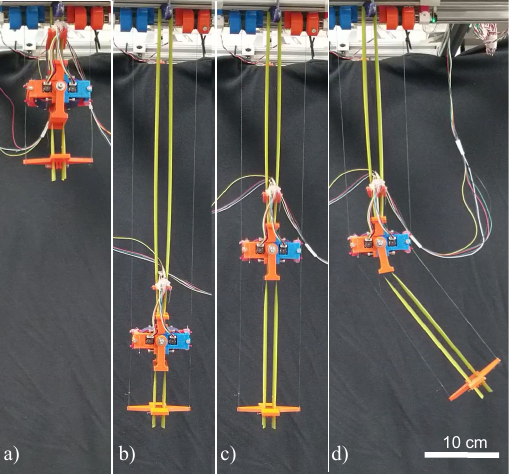}
    \caption{Snapshots in the actuation sequence of the 3-DOF pinched tape manipulator. a) Initially, the arm can be stored compactly before b) lengthening by extending the tapes. c) Afterwards, the pinching node can be moved along the tape to set the bend point. d) An in-plane bend can then be created by pulling on one of the cables. The design is lightweight and has large extension compared to traditional robot manipulators.}
    \label{fig:CoreFunctionality}
\vspace{-10pt}
\end{figure}

Rather than longitudinally bending the tape to engage the tape's bistability and produce a change in direction, we propose pinching the tape to flatten its transverse curvature, reducing its bending stiffness to produce a region where localized bending can occur. This localized bend point functions as a revolute joint. Prior work used the longitudinal curvature to produce a hinge~\cite{Vehar2004CLOSEDLOOPTS}, whereas we use the change in transverse curvature.

In related work, researchers have previously investigated varying geometric parameters to induce stiffness change for bending in pressurized tubes~\cite{Gravish2021VacuumPinch, do2020dynamically}. Prior work has also investigated shape change through programmable stiffness at discrete joints via material phase change and shape memory polymers~\cite{Firouzeh2017, McEvoy2016}. However, the topic of continuously reconfigurable joints has been less explored~\cite{Usevitch2020Truss}.

Figure~\ref{fig:CoreFunctionality} shows our lightweight, high-extension, planar 3-degree-of-freedom (DOF) cable-driven manipulator. This demonstrates a novel, compact design capable of reconfigurable bending. To our knowledge, tape springs have not previously been used as the basis of a bending retractable manipulator. Section II presents the overall design concept and implementation of the manipulator. Section III models the change in joint stiffness and the kinematics and workspace of the robot. Section IV shows demonstrations that illustrate the manipulator's capabilities, and Section V concludes with the implications of this system and areas for future work.

\section{Design and Implementation}
\begin{figure}
\vspace{1.5mm}
    \centering
    \includegraphics[width=\linewidth]{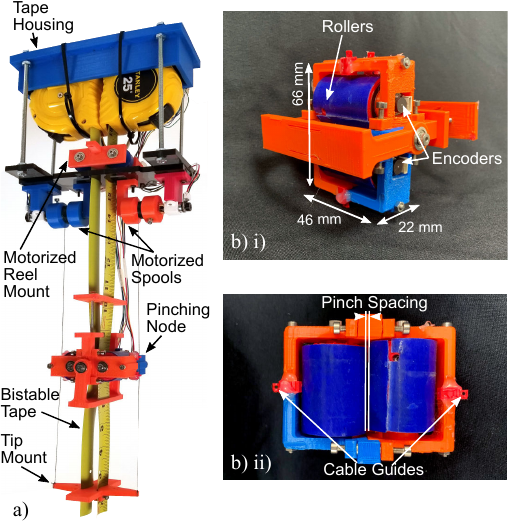}
    \caption{a) The overall system, including two bistable tapes connected at a base, pinching node, and tip mount. b) The pinching node consists of (i) two rollers which (ii) pinch the tapes to create a localized bend.}
    \label{fig:FigureConcept}
\end{figure}

Our manipulator is a cable-driven 3-DOF serial planar manipulator consisting of two bistable tapes arranged back-to-back, as shown in Fig.~\ref{fig:FigureConcept}. A revolute joint is formed through pinching the tapes together. The relative link lengths are determined through the overall tape lengths, which spool from the base, and the position of the pinching node. The angle of the two links with respect to each other is determined by the lengths of two cables which travel from the base to the node to the tip, on opposite sides of the manipulator. 

The design consists of three main components: 1) the bistable tapes, which form the backbone of the manipulator and are stored in reels at the base; 2) a pinching node, which traverses the tape and whose rollers create a pinch point that functions as a revolute joint; and 3) two sets of motorized spools, one set which controls the cables which enable steering and the other set which extend/retract the tape from/into their spools.

The weight of the system without the tape measures is 535~g. Of that, the pinching node weighs 163~g and the tape housing, reel mount, and spools comprise the remaining 372~g. The off-the-shelf tape measures (Stanley) each have a 7.62~m (25~foot) long tape and weigh 368~g. Depending on the extension length desired, much of this weight could be removed -- the weight of 3~m (9.8~feet) of tape is only 76~g. Furthermore, there is redundancy in the current housing design of the tape measures, as we secure the individual tape measure plastic casings into a larger 3-D printed housing. Thus, future designs could be an even lighter weight, with the bistable tapes being directly wound onto the motorized reel mounts and stored in a single housing. 

\subsubsection{Bistable Tape}
Two bistable tapes form the backbone of the manipulator and are stored in reels at the base. The two tapes are placed back-to-back, resulting in symmetric bending behavior as well as increasing the overall stiffness. The stiffness of a bistable tape is dependent on its transverse curvature. We used pre-stressed 0.2~mm thick bistable steel tapes from commercial off-the-shelf tape measures for our manipulator. These tapes possess a constant positive transverse curvature which is imparted to the steel strips via heat treatment and plastic bending during manufacturing. This also gives the tapes their bistability. The tapes can exist extended, with zero longitudinal curvature and non-zero transverse curvature, or coiled, with non-zero longitudinal curvature and zero transverse curvature; this latter property enables the tapes to be stored compactly in spools. 

When the tape is extended, its transverse curvature imparts longitudinal bending stiffness. The manipulator backbone is formed by placing two tapes back-to-back, ensuring that one tape experiences opposite sense bending regardless of bend direction and resulting in increased arm stiffness. The effect of transverse curvature on bending stiffness is discussed in detail in Sec.~\ref{Sec Stiffness}.

\subsubsection{Pinching Node}
The pinching node uses a set of motorized rollers. The width between the two rollers is predetermined to pinch the two tapes against one another, flattening them so that they have zero transverse curvature along the contact area with the rollers. Therefore, each tape is passively pinched by the rollers.

The tape acts as a linear track for the pinching node. When the rollers are not active, the node is simply clamped in place on the tape. Extension/retraction of the tape does not change the node's position relative to the tape. When active, the rollers are used to drive the node along the tape.

Each roller is wrapped in a high-friction non-slip material, Dycem (Dycem Corporation), to ensure good contact with the tape and is controlled by a micro metal gearmotor (Pololu) located inside the roller; the roller is secured to the motor shaft via a set screw. Both motors have encoders to measure their rotation. The rollers are housed in a 3-D printed frame, which contains cable guides that route cables through the node (between the base and the tip mount) and also attaches to a hinge, which rotates about the pinching point and allows the tapes to exit after being pinched. The hinge has joint limits of $\pm55^\circ$, beyond which the hinge interferes with the main node body.

Above and below the pinch point, the frame forces the tape to return back to its original separation distance. Thus, the pinch results in a localized change in curvature and tape separation distance rather than a gradual tapering. In the absence of cable tensions, the joint does not affect the tape shape and the tape remains straight. This is unlike examples of tape spring joints in the literature~\cite{Vehar2004CLOSEDLOOPTS}, which result in larger global curvature of the tape. 

\subsubsection{Motorized Spools for Cables and Tapes}
In addition to the pinching node, there are two additional sets of motorized rollers: one for controlling the tape extension/retraction and the other for pulling the left and right cables to steer the manipulator.

The reel mount contains two motorized rollers -- one for each tape -- wrapped in Dycem. The two work together to reel both tapes at the same rate. Above the reel mount are two motorized pulleys that spool the cables. These cables run from the base, through cable guides in the sides of the pinching node, and are secured at the tip mount. Thus, forces from the cable tension are transmitted to the end effector. A bend is formed by shortening one of the cables. Once a bend is formed, its joint stiffness is determined by the cable tensions.

\section{Modeling}
In this section, we describe the change in stiffness due to curvature change from the pinching node, which results in the creation of a movable joint, and derive the manipulator kinematics.

\subsection{Stiffness Change Due to Pinching} \label{Sec Stiffness}

\begin{figure}
\vspace{1.5mm}
    \centering
    \includegraphics[width=\linewidth]{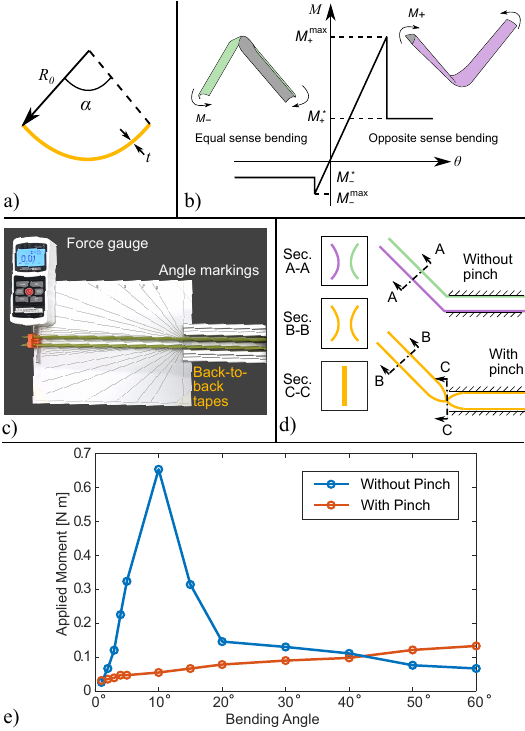}
    \caption{Bending of bistable tapes. a) Cross section diagram of unstressed bistable tape. b) Conceptual schematic of moment vs. bending angle for a single bistable tape in equal sense bending (green) and opposite-sense bending (purple)~\cite{Pellegrino1999}. c) Experimental setup to measure moment vs. bending angle for back-to-back tapes. d) Diagram with overhead and section views of bending for unpinched and pinched tapes. e) Measured moment vs. bending angle for back-to-back bistable tapes with and without a pinch. Bending angle stops at $60^\circ$ because the test conditions leave the configuration space of the full manipulator after that angle. }
    \label{fig:Stiffness}
\end{figure}

The bending stiffness of a bistable tape is determined by its geometry and material parameters. A bistable tape is characterized by its transverse curvature, which is defined by its unstressed radius of curvature $R_0$ and subtended angle $\alpha$, as shown in Fig.~\ref{fig:Stiffness}a).

In the case where the tape is pinched, it becomes a flat strip with a rectangular cross section rather than the curved section seen in Fig.~\ref{fig:Stiffness}a). Therefore, the required applied bending moment $M$ of the flattened tape is:
\begin{equation}
    M = EI\kappa = ER_0\alpha \frac{t^3}{12}\kappa
\end{equation}
\noindent where $E$ is the elastic modulus of the tape material, $I$ is the second moment of area of the cross section, $t$ is the tape thickness, and $\kappa$ is the longitudinal beam curvature.

In the case where the tape is not pinched, bending stiffness depends on the direction in which the tape is bent. Figure~\ref{fig:Stiffness}b) shows a conceptual schematic of the applied moment versus bending angle for equal and opposite sense bending of a single tape~\cite{Pellegrino1999}. If the tape is bent in-plane such that the external torque is in the same direction as the transverse curvature, the tape experiences equal sense bending. If the external torque is opposite to the transverse curvature, the tape experiences opposite sense bending. In equal sense bending, the tape exhibits low stiffness and quickly transitions to a state of uniform longitudinal curvature when subjected to a moment. In contrast, the tape is much stiffer in opposite sense bending, with a high stiffness for small rotation angles followed by buckling and a dramatic drop in bending stiffness with the formation of a fold. 






In the pinched tape arm, the tapes are oriented back-to-back, such that their transverse curvatures are oriented in opposite directions. In the arm, the tape serves as both a structural element to support loads as well as the medium through which bends are produced. To gain insight into how locally changing transverse curvature affects bending stiffness for this arrangement, we conducted tests in which we measured the applied moment versus bending angle for back-to-back bistable tapes, with and without the presence of a pinching point.

Figure~\ref{fig:Stiffness}(c) shows the experimental setup. For this test, two steel bistable tapes were secured into 3-D printed mounts at both ends in order to fix the distance between the two tapes. One mount was secured to prevent rotation while the other end was affixed to a Mark-10 Series 5 force gauge, which was mounted orthogonal to the tape length. Using printed markings, the tape was bent to a specified angle and the applied moment was recorded. For the pinched tape tests, a 3-D printed slit was used to pinch the tapes together. Figure~\ref{fig:Stiffness}(d) shows a schematic of the back-to-back tapes in this experiment.

 Figure~\ref{fig:Stiffness}e) shows the result of these measurements. The applied moment exhibits a strong nonlinearity for the case without a pinch. The maximum applied moment to bend the non-pinched tapes is also greater, reaching 0.654~N$\cdot$m, whereas at that same angle, the moment was 0.055~N$\cdot$m for the pinched tapes. We observed that this behavior was symmetric for positive and negative bending angles due to the back-to-back tape configuration.
 
 In the non-pinched case, as illustrated in Fig.~\ref{fig:Stiffness}d), one of the tapes in the back-to-back setup experiences opposite sense bending while the other experiences equal-sense bending at the bending point. Whereas in the pinched case, we avoid experiencing $M^\text{max}_-$ and $M^\text{max}_+$ because the tapes are flat at the bending point. This represents a key difference between bends produced with and without pinching, the latter of which is the mode seen in prior work.
 
Thus, by pinching the tapes and locally setting the transverse curvature to zero, we can eliminate the snap-through instability otherwise experienced when producing bends in bistable tapes. By avoiding this buckling, we can control small bending angles and seamlessly transition to larger bending angles with our manipulator. Furthermore, we can precisely control where bending will occur because the tape preferentially bends at the pinching point produced by the rollers in our pinching node.
 

\subsection{Manipulator Kinematics} \label{Section Kinematics}

\begin{figure}
\vspace{1.5mm}
    \centering
    \includegraphics[width=\linewidth]{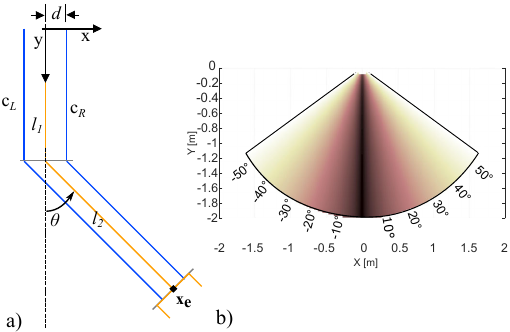}
    \caption{a) Pinched tape manipulator kinematics. b) Manipulator workspace plot with contour lines showing the minimum possible angle at which a point $(x,y)$ can be reached. The workspace is symmetric about $x=0$. Points that are at an angle greater than $\pm55^\circ$ from the midline are outside the workspace. Darker shading corresponds with a greater number of configurations able to reach the specified end effector position.}
    \label{fig:Kinematics}
\end{figure}

The pinched bistable tape manipulator is a serial-chain prismatic-revolute-prismatic (PRP) planar manipulator with coupling between link lengths. Figure~\ref{fig:Kinematics}a) shows the parameters that define the robot kinematics. 

For link lengths $\ell_1$, $\ell_2$ and bending angle $\theta$, the forward kinematics for the end effector $\textbf{x}_e$ are described by:
\begin{equation}
    \begin{bmatrix} x \\ y\end{bmatrix}   = \begin{bmatrix}  0 & \sin\theta \\ 1 & \cos\theta \end{bmatrix} \begin{bmatrix} \ell_1 \\ \ell_2 \end{bmatrix}
    \label{eq base PRP manipulator}
\end{equation}

We define the bending angle $\theta$ as the angle from the midline of the base to the midline between the tapes in Link 2. $\theta$ can be set independent of the lengths $\ell_1, \ell_2$ and is determined by the relative lengths $c_L, c_R$ of the left and right cables, respectively. For a fixed cable offset $d$ from the midline of the manipulator, the cable lengths are given by:
\begin{equation}\label{eq cable lengths}
\begin{split}
    c_L &= \ell_1 + \ell_2 + d\sin\theta \\
    c_R &= \ell_1 + \ell_2 - d\sin\theta
\end{split}
\end{equation}

The lengths $\ell_1, \ell_2$ are determined by two control parameters: the growth of the bistable tapes as a result of extension/retraction from the base, $q_1(t)$, and the position of the node, $q_2(t)$. Extension of the tape results in an increase in the total length $L = \ell_1 + \ell_2$ through an increase in $\ell_1$. Meanwhile, $\ell_2$ is not affected by growth from the base. 

The position of the node $q_2(t)$ defines $\ell_1$ and is limited to $0\leq n \leq L$. A change of $\dot{q}_2$ for the node, results in a corresponding change of $\dot{q}_2$ in the length $\ell_1$ and a change of $-\dot{q}_2$ in the length $\ell_2$; thus, $L$ remains constant. Therefore, the link lengths $\ell_1(t)$ and $\ell_2(t)$ are given by:
\begin{equation}
    \begin{bmatrix} \ell_1 \\ \ell_2\end{bmatrix}   = \begin{bmatrix}  1 & 1 \\ 0 & -1 \end{bmatrix} \begin{bmatrix} q_1 \\ q_2 \end{bmatrix} + \begin{bmatrix} \ell_1(0) \\\ell_2(0) \end{bmatrix}
    \label{eq growth and node}
\end{equation}

\noindent where $\ell_1(0)$ and $\ell_2(0)$ are the link lengths at time $t=0$ of Links 1 and 2, respectively. Combining Eqs.~\ref{eq base PRP manipulator} and \ref{eq growth and node}, we obtain:
\begin{equation}
    \begin{bmatrix} x \\ y\end{bmatrix}   = \begin{bmatrix}  0 & -\sin\theta \\ 1 & 1-\cos\theta \end{bmatrix} \begin{bmatrix} q_1 \\ q_2 \end{bmatrix} +  \begin{bmatrix}  0 & \sin\theta \\ 1 & \cos\theta \end{bmatrix} \begin{bmatrix} \ell_1(0) \\\ell_2(0) \end{bmatrix}
    \label{eq overall kinematics}
\end{equation}

\subsection{Workspace}
Using the kinematics derived in Sec.~\ref{Section Kinematics}, we can calculate the workspace of the pinched tape manipulator. Figure~\ref{fig:Kinematics}b) shows the workspace for a manipulator based at $(0,0)$ and a maximum total link length of 2~m. In general, the workspace limits are set by the maximum bending angle of the pinching and the total available tape length. Thus, ideally, for the hinge rotation limits of $\pm55^\circ$, all points that are within $\pm55^\circ$ from the y-axis and a radius of 2~m from the origin are in the workspace, as shown in Fig.~\ref{fig:Kinematics}b). In reality, for this manipulator, there is a minimum link length associated with $\ell_1$ of 7.6~cm, which corresponds to the length of the pinching node. The tape must be at least 7.6~cm long for it to be able to exit the pinching node. As points approach the joint limit boundaries, the minimum end effector angle increases and approaches $55^\circ$. For points within the workspace, the end effector can reach with orientations above the minimum end effector angles, except along $x=0$ where the only feasible orientation is $0^\circ$.

\section{Demonstrations and Experiments}

Figure~\ref{fig:CoreFunctionality} illustrates the three control inputs of the bistable tape manipulator performed sequentially: 1) length change of the tape, 2) movement of the pinching node along the tape, and 3) tensioning of the individual cables and bending of the overall manipulator.

These inputs can be performed simultaneously, resulting in full control of the end effector planar position and orientation. We conducted a series of demonstrations to further validate and illustrate the manipulator's capabilities.

\begin{figure}[b]
    \centering
    \includegraphics[width=\linewidth]{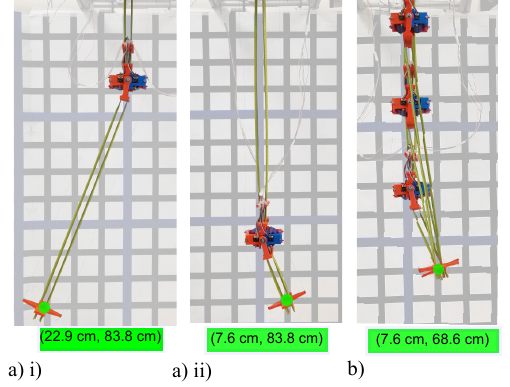}
    \caption{a) Varying node position to access different section lengths (i) short $\ell_1$ and long $\ell_2$, (ii) long $\ell_1$ and short $\ell_2$. b) Overlay of multiple manipulator configurations maintaining the same end effector position }
    \label{fig:MultipleShapes}
\end{figure}

 \subsection{Reaching Demonstrations}
 Two common tasks required of a planar manipulator are to reach multiple targets in space and to reach the same target with different orientations/configurations. We conduct demonstrations that illustrate each using the kinematics derived in Sec.~\ref{Section Kinematics}.
  
 \subsubsection{Targets in free space}
 For our first set of tests, we moved the end effector to specified targets. The reconfigurable nature of the bend position allows us to produce configurations with a long $\ell_{1,2}$ and/or a short $\ell_{1,2}$. This capability can allow for more flexibility in task planning for the manipulator during deployment. For example, we can envision the benefits of having a configuration like that in Fig.~\ref{fig:MultipleShapes}a) i) where $\ell_2$ can extend as required to tackle tasks in which the object to be grasped is recessed in a confined space, such as within a small crevice.
 
 Figure~\ref{fig:MultipleShapes}a) shows two examples of the manipulator end effector successfully reaching specified target positions. For the examples shown in Fig.~\ref{fig:MultipleShapes}a), we verified the end effector position using a 3~in by 3~in grid. We commanded the end effector to coordinates (22.9 cm, 83.8 cm) and (7.6 cm, 83.8 cm) for i) and ii), respectively.
 
 
 \begin{figure}[b]
    \centering
    \includegraphics[width=\linewidth]{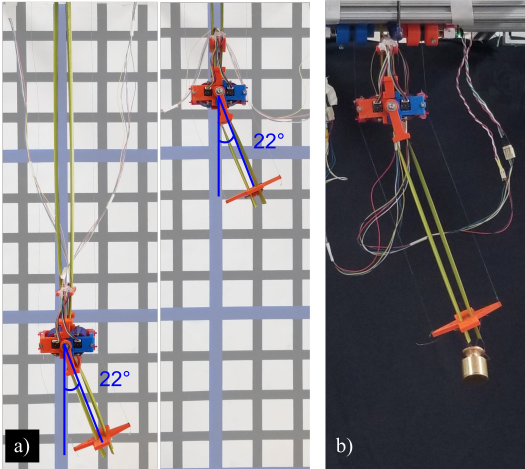}
    \caption{a) The same end effector orientation can be maintained while retracting the tapes. b) Here the arm has lifted up a 200~g weight at a specified angle.}
    \label{fig:SameL2Angle}
\end{figure}

\begin{figure}
\vspace{1.5mm}
    \centering
    \includegraphics[width=\linewidth]{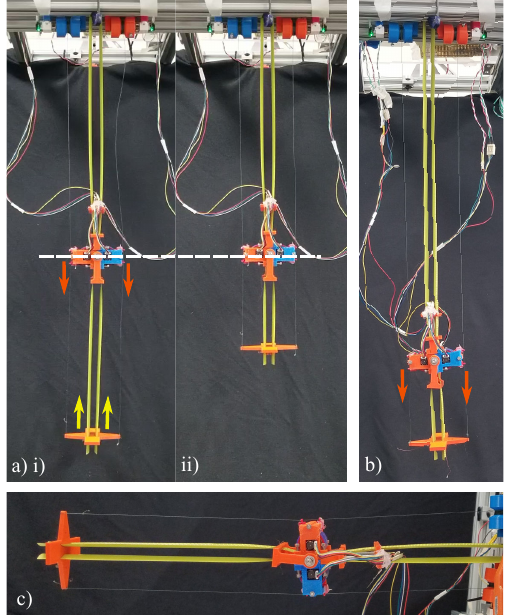}
    \caption{a) The pinching node can be moved at the same rate at which tape is grown, resulting in a stationary bending point. The position of the node is shown by the white dotted line and is the same in configurations (i) and (ii). b) Without coordinating the rates of tape growth and pinching node displacement, the ability to vary the length of $\ell_2$ without changing $\ell_1$ is lost. c) The arm can be mounted at different orientations. Here, the arm is mounted horizontally and cantilevered over free space.}
    \label{fig:OtherCapabilities}
\end{figure}
 
 \subsubsection{Multiple configurations to reach the same target}
 Being able to reach a target at a desired orientation or having multiple configurations to reach a desired target can both increase the utility of a manipulator. To that end, we conducted experiments where we use multiple configurations to reach a specified target.
 
 Figure~\ref{fig:MultipleShapes}b) shows, using 3 different overlaid configurations, that the end effector orientation and link lengths can be varied to maintain a static end effector position. For this test, we chose an arbitrary point within the manipulator workspace. The end effector is first commanded to (7.6~cm, 68.6~cm) in the first configuration which has a joint angle of $7.1^\circ$. In this configuration, $\ell_1$ and $\ell_2$ are at lengths of 7.6 cm and 61.5 cm, respectively. Then, the node position is adjusted to the second configuration while commanding the end effector to maintain its position. With this placement, the joint angle becomes $10^\circ$ and $\ell_1$ and $\ell_2$ change to 25.4 cm and 43.8 cm, respectively. Finally, the node position is adjusted to the third configuration with the desired target staying constant. The joint angle then becomes $16.7^\circ$ with $\ell_1$ resting at 43.2 cm and $\ell_2$ at 26.5 cm. 
 
 \subsection{Bending Demonstrations}
 \subsubsection{Constant end effector orientation during length change}
 The control of bend angle $\theta$ requires control of the cable lengths. When the manipulator lengthens due to extension of the tapes, maintaining a constant end effector angle thus requires also changing the cable lengths.
 
 We demonstrate this capability in Fig.~\ref{fig:SameL2Angle}a). Here, as the tape is retracted, we control the cable tensions such that the angle $\theta$ remains constant at $22^\circ$ during retraction. Maintaining a constant end effector angle would allow grasped objects to be held in the same orientation as they are pulled towards the aerial robot body for more secure in-flight transport. Figure~\ref{fig:SameL2Angle}b) shows the arm  lifting up a 200~g weight to a specified angle.
 
 
 \subsubsection{Constant Link 1 length while shortening Link 2}
 
 A key difference between the pinched tape manipulator and a traditional PRP manipulator is the coupling of the link lengths. In a traditional PRP manipulator, the lengths of Links 1 and 2 can be varied independently of the other. For example, for a conventional PRP robot, Link 2 can be shortened while Link 1 remains constant. However, as Eq.~\ref{eq growth and node} shows, the control inputs for $\ell_1$ and $\ell_2$ are coupled.
 
For the pinched tape manipulator to function with the same kinematics as a PRP manipulator, it is thus necessary to be able to control the rate of growth/retraction and movement of the pinching node to maintain a bending point that remains stationary with respect to an external frame. This can be done by matching the magnitudes of the rates of growth/retraction and pinching node traversal.

We demonstrate this capability in Fig.~\ref{fig:OtherCapabilities}a). Figure~\ref{fig:OtherCapabilities}a) shows two snapshots side-by-side of the manipulator as the tape is retracted into the base. By driving the node forward and matching the magnitudes of $\dot{q}_1$ and $\dot{q}_2$, the pinching node keeps $\ell_1$ constant. If Link 2 was shortened solely through driving the node towards the tip, it would also result in lengthening of Link 1, as seen in Fig.~\ref{fig:OtherCapabilities}b).

Additionally, we tested this same process for arms with a bend and also validated the ability to successfully maintain a desired bend angle while shortening $\ell_2$. 
 
 
 \subsubsection{Extension and mounting}
 
For aerial manipulation, the tape arm is ideally mounted pointing downwards. However, in general, our design can be mounted to point in other directions. Figure~\ref{fig:OtherCapabilities}c) shows the arm mounted horizontally and cantilevered over free space. In general, we can mount at arbitrary angles -- we have successfully grown and formed bends when the manipulator is pointing downwards, pointing upwards, and cantilevered parallel to the ground.
 
 The arm is capable of extending to large lengths, particularly when deployed pointing downwards, where the tapes can be loaded in tension. This case is analogous to the tether-supported payloads. In such cases, the only limit to the extension is the amount of material that can be spooled. By fully unspooling the spooled bistable tapes, we can achieve extension ratios in excess of 20:1. Even when bent, the arm still achieves large extensions, with the limiting factor now determined by the buckling failure mode of the tapes.

\section{Conclusion and Future Work}
We present the novel concept and design of a reconfigurable manipulator using pinched bistable tapes to allow for a lightweight, compact robot arm capable of high extension. We describe the bending stiffness and manipulator kinematics in addition to demonstrating use of the manipulator. 

A number of directions exist for future work. Our long-term goal is to build upon the presented manipulator and incorporate it with a aerial robot for more capable aerial manipulation. Mounting the manipulator onto an aerial robot will offer additional degrees of freedom and broaden the feasible workspace. One concrete improvement to reduce weight is to refine the design of the housing to more compactly spool the bistable tape around the motorized reels. A gripper on the end of the arm would allow for grasping and object manipulation, and sensors such as cameras would allow the device to be used for inspection. We will further investigate possible failure modes for the arm and determine how the arm's structure and configuration affects its maximum payload.

We also plan to further characterize and investigate dynamic control of the manipulator. In general, this bistable tape manipulator with its low cost, low weight, compliance, and dexterity can be thought of as intermediate between passive, cable-suspended loads and complex, traditional rigid robot arms; the bistable tape manipulator is capable of acting in ways similar to both. For example, it may be possible to use the manipulator to swing loads akin to a tether. Investigation of the dynamics and stiffness control could thus enable a broader array of aerial manipulation tasks.





\bibliographystyle{IEEEtran}
\bibliography{references}
\end{document}